\documentclass[10pt,twocolumn,letterpaper]{article}

\usepackage{cvpr}              % To produce the CAMERA-READY version
\usepackage{graphicx}
\usepackage{amsmath}
\usepackage{amssymb}
\usepackage{booktabs}
\usepackage{multirow}
\usepackage{tabularx}
\usepackage{float}
\usepackage{listings}

\usepackage[pagebackref,breaklinks,colorlinks]{hyperref}

\usepackage[capitalize]{cleveref}
\crefname{section}{Sec.}{Secs.}
\Crefname{section}{Section}{Sections}
\Crefname{table}{Table}{Tables}
\crefname{table}{Tab.}{Tabs.}

\def\cvprPaperID{*****} % *** Enter the CVPR Paper ID here
\def\confName{CVPR}
\def\confYear{2025}

\begin{document}

\title{Interpreting and Mitigating Hallucination in MLLMs through Multi-agent Debate}

\author{Zheng Lin\\
Xidian University\\
{\tt\small linzheng@stu.xidian.edu.cn}
\and
Zhenxing Niu\\
Xidian University\\
{\tt\small zxniu@xidian.edu.cn}
\and
Zhibin Wang\\
INF Tech Co., Ltd.\\
{\tt\small zhibin.waz@inftech.ai}
\and
Yinghui Xu\\
INF Tech Co., Ltd.\\
{\tt\small xuyinghui@inftech.ai}
}
 
\maketitle

%%%%%%%%% ABSTRACT
\begin{abstract}
MLLMs often generate outputs that are inconsistent with the visual content, a challenge known as hallucination. Previous methods focus on determining whether a generated output is hallucinated, without identifying which image region leads to the hallucination or interpreting why such hallucinations occur. In this paper, we argue that hallucination in MLLMs is partially due to a lack of slow-thinking and 
divergent-thinking in these models. To address this, we propose adopting a self-reflection scheme to promote slow-thinking. Furthermore, we consider eliminating hallucination as a complex reasoning task and propose a multi-agent debate approach to encourage divergent-thinking. Consequently, our approach can not only mitigate hallucinations but also interpret why they occur and detail the specifics of hallucination. In addition, we propose to distinguish creativity from hallucination in the context of MLLMs, and illustrate how to evaluate MLLMs' creativity capability. Extensive experiments on various benchmarks demonstrate that our approach exhibits generalized hallucinations-mitigating performance across several MLLMs. The code is available \href{https://github.com/LZzz2000/MAD}{here}.
\end{abstract}

%%%%%%%%% BODY TEXT
\section{Introduction}
\label{sec:intro}
Recently, large language models (LLMs) have made remarkable strides, dominating a broad spectrum of tasks in natural language processing (NLP) such as language comprehension~\cite{hendrycks2021}, generation~\cite{zhang2023benchmarking}, and reasoning~\cite{chu2024naviga, qiao2023reasoning}. Consequently, research focus has increasingly shifted towards multimodal large language models (MLLMs), also known as large vision-language models (LVLMs).
These MLLMs have demonstrated promising capabilities in multimodal tasks such as image captioning~\cite{li2023blip2}, visual question answering~\cite{dai2023instructb}, \emph{etc}. However, the issue of hallucination within LLMs and MLLMs has become a significant research focus, as it poses a challenge to their reliability and applicability.

\begin{figure}[tbp]
\centering
\subfloat[``Is there a fork?" -- ``Yes."]{
\includegraphics[width=0.34\linewidth]{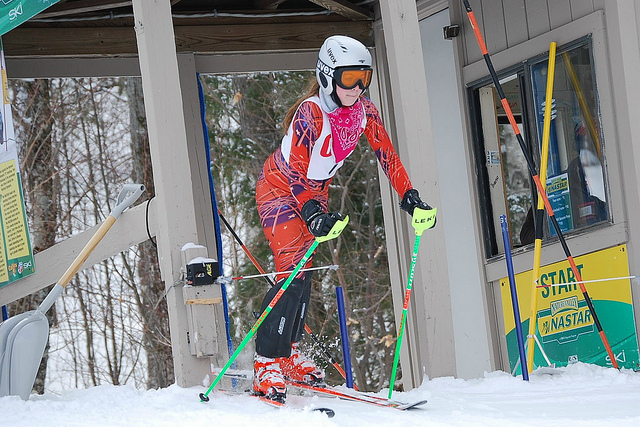}
}
% \subfloat[``Is there a train?" -- ``Yes."]{
% \includegraphics[width=0.34\linewidth]{pic/interpretation/COCO_val2014_000000091954.jpg}
% }
\subfloat[``Is there a bowl?" -- ``Yes."]{
\includegraphics[width=0.34\linewidth]{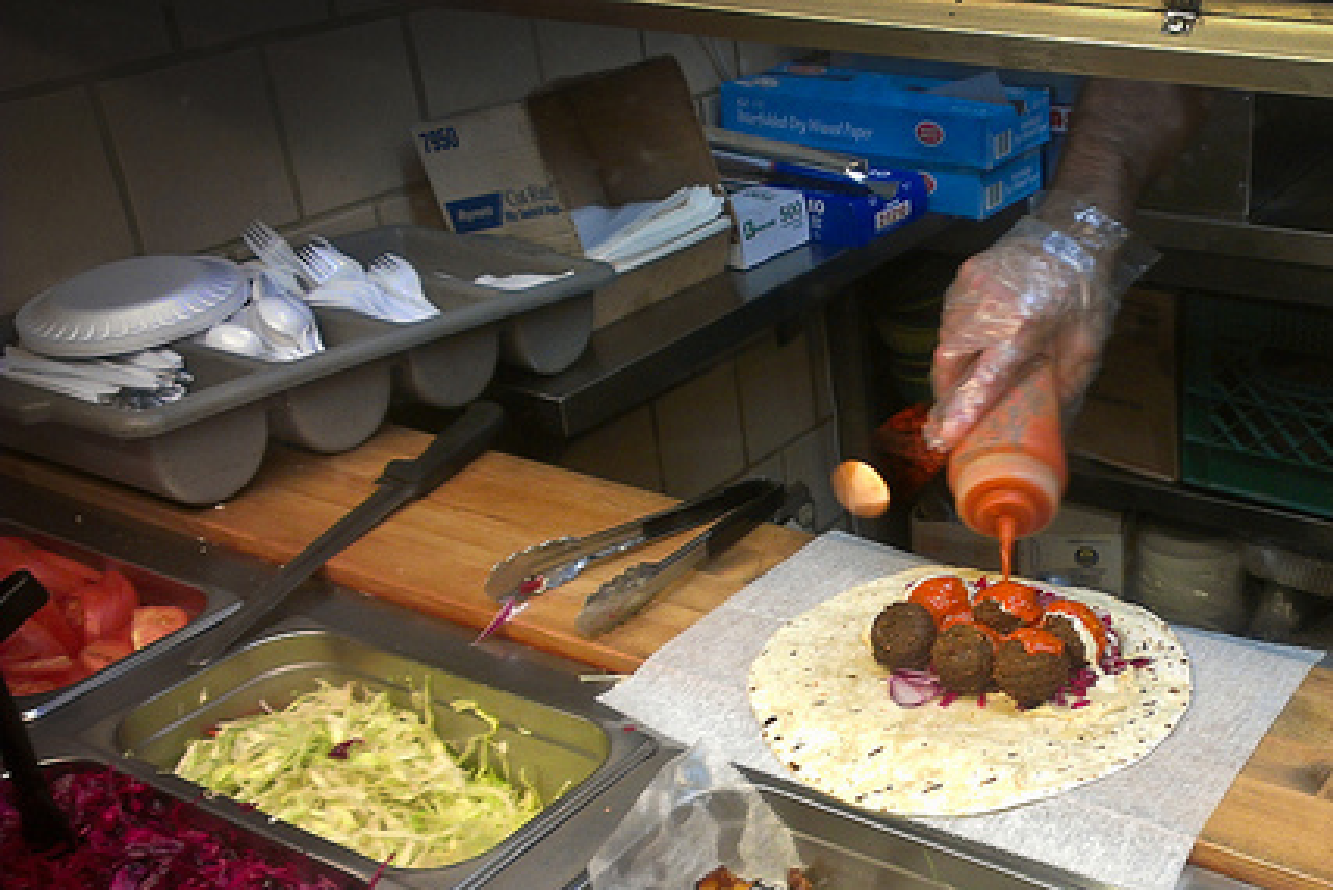}
}
\subfloat[``Is there a backpack?" -- ``Yes."]{
\includegraphics[width=0.34\linewidth]{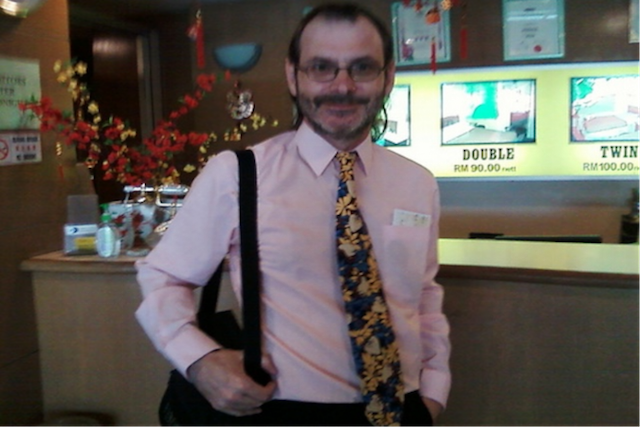}
}
\\
\subfloat[``Is there an orange?" -- ``Yes, because the frisbee is orange."]{
\includegraphics[width=0.34\linewidth]{pic/object-ambiguity/orange-original-image.jpg}
}
\subfloat[``Is there a bed?" -- ``Yes, because there is a mattress."]{
\includegraphics[width=0.34\linewidth]{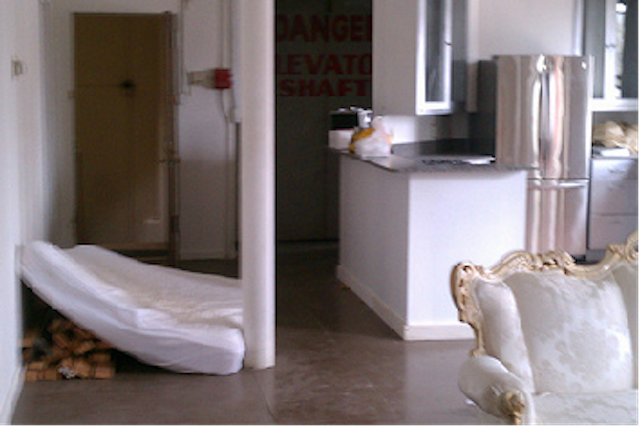}
}
\subfloat[``Is there a laptop?" -- ``Yes, because there is a keyboard."]{
\includegraphics[width=0.34\linewidth]{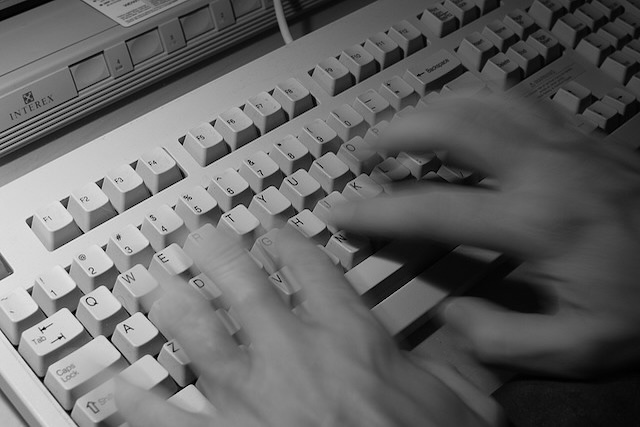
}}
\caption{Our approach can interpret the causes of an MLLM hallucination: (a) MLLM mistakenly identified a ``ski pole" as a ``fork" due to \emph{limited} perceptual ability; (b) MLLM mistakenly identified a ``plate" as a ``bowl" due to their \emph{visual} similarity; (c) MLLM mistakenly identified a ``handbag" as a ``backpack" due to their \emph{conceptual} similarity; (d) MLLM \emph{misunderstands} the ``orange" in the question as a color; (e) MLLM mistakenly \emph{inferred} the existence of a ``bed" upon perceiving a ``mattress"; (f) MLLM mistakenly \emph{inferred} the existence of a ``laptop" upon perceiving a ``keyboard". Note that (e) and (f) are due to unreasonable guesses/inferences.}\label{fig1}
\vspace{-1.0em}
\end{figure}

Hallucination in LLMs typically manifests as a discrepancy between generated content and actual facts, known as \emph{factual hallucination}~\cite{bai2024}. In contrast, hallucination in MLLMs often appears as a mismatch between the generated text response and the provided visual content, referred to as \emph{cross-modal inconsistency}~\cite{bai2024}. Compared to LLM hallucination, MLLM hallucination tends to be related to the detailed content of images, making it easier to occur and harder to identify and mitigate. Therefore, numerous approaches have been proposed to mitigate hallucinations in MLLMs~\cite{sun2023, yu2024hallucidoctor}. Current mainstream solutions include leveraging hallucination-targeted datasets for fine-tuning, training a post-hoc revisor to produce less hallucinatory outputs, and adapting factually augmented Reinforcement Learning from Human Feedback (RLHF).

While these methods demonstrate commendable performance, %they incur substantial additional costs, such as the annotation budget for additional instruction data~\cite{yu2024hallucidoctor}. More importantly, these methods 
they fail to interpret why these hallucinations occur. Particularly, they only determine whether a generated output is hallucinated or not, without providing the evidence for this judgment or finding out which image region leads to the hallucination. We argue that identifying the causes of MLLMs' hallucinations can significantly aid in their mitigation.
As shown in Fig.\ref{fig1}, hallucinations may arise due to distinct reasons, such as the limited ability to perceive image content, misunderstanding of concepts in the questions, tendency to make unreasonable/excessive associations, \emph{etc}. Thus, we believe that the \emph{interpretation} of hallucination is desirable, as it aids not only in mitigating hallucinations but also in pointing out the direction to improve the \emph{overall} capability of MLLMs.

%lack the capacity for slow-thinking, and do not clearly distinguish creativity from hallucinations.

On the other hand, recent studies have demonstrated that LLMs still struggle with reasoning-intensive tasks. Generally, complex reasoning tasks often require cognitive capabilities, such as \emph{slow-thinking}~\cite{chen2024slow} and \emph{divergent-thinking}~\cite{liang2024divergentthink}. Therefore, 
many methods have been proposed to enhance LLMs' cognitive capabilities~\cite{chan2023chateval}. Self-Reflection (SR) strategy~\cite{shinn2023} has been proposed to improve LLM's capability on slow-thinking, which involves an iterative refinement process such that the LLM generates a new answer based on the answers and feedback in previous iterations. To enhance the divergent-thinking capability, the ``society of minds" approach (also known as \emph{cognitive synergy} approach)~\cite{CognitiveSynergy, chan2023chateval} has been proposed, which emphasizes collaboration among different individuals. An representative strategy is the Multi-Agent Debate (MAD)~\cite{liang2024divergentthink}. In this strategy, multiple language model instances (or agents) individually propose and jointly debate their responses and reasoning processes to arrive at a common answer.

Inspired by this, we argue that eliminating hallucination can be regarded as a complex reasoning task, requiring us to identify the cause of the hallucination and provide a detailed interpretation. Therefore, in this paper, we propose leveraging self-reflection and multi-agent debate strategies to mitigate hallucination in MLLMs.
Firstly, we adopt the self-reflection strategy by asking a MLLM a \emph{series} of questions (\emph{i.e.,} a sequence of ``question, answer, new question, new answer, ...") related to the hallucinated object's color, shape, spatial location, relation to other objects, \emph{etc}. This \emph{continuous inquiry} promotes slow-thinking for MLLMs. 

Secondly, we adopt the multi-agent debate strategy by instantiating several MLLM as agents to conduct a multi-round debate. Specifically, when \emph{Agent A} generates its answer, we will ask \emph{Agent B} to analyze A's answer and provide its judgment. Then, Agent A will consider B's answer and reevaluate its previous answer. After several rounds of debate, if they cannot reach an agreement, we will ask the \emph{Agent Judge} to give a conclusion.
This cross-agent debate encourages divergent-thinking and promotes cognitive synergy. Additionally, after the debate procedure, we can not only identify hallucinations but also interpret why they occur and detail the hallucination's specifics, \emph{e.g.,} which image region caused the hallucinated descriptions, and what the actual object that region is, as shown in Fig.\ref{fig3}.

Regarding multi-agent debate, it is crucial to adopt a workflow tailored to the specific task. For hallucination mitigation, we have designed a specific debate workflow and crafted appropriate prompts for all debaters. Specifically, we found it is essential to: (1) assign distinct personalities (\emph{e.g.,} conservative or imaginative) to different debaters; (2) propagate partial rather than full information among debaters; (3) pose questions strategically; and (4) leverage the imitation learning techniques.
An overview of the debate workflow is illustrated in Fig.\ref{fig2}.
It is worth noting that our approach leverages MLLM as agents in a black-box manner, \emph{i.e.,} by calling the MLLM's API without requiring knowledge of the model's architecture and parameters. 

Additionally, we corrected some annotation errors in the POPE dataset, ensuring a precise evaluation of hallucination mitigation. Furthermore, recent research has suggested distinguishing creativity from hallucination in LLMs. Inspired by this, we propose to distinguish creativity from hallucination in the context of MLLMs and propose a dataset to evaluate MLLMs' capability on creativity.

%-------------------------------------------------------------------------

%With extensive experiments on various benchmarks and hallucination metrics, our approach exhibits generalized hallucinations-reducing performance on several MLLMs. Our contributions can be summarized as follows:
%\begin{itemize}
    %\item We argue that hallucination in MLLMs is partially due to a lack of slow-thinking and propose adopting a self-reflection scheme to address this issue. 
    %\item We consider eliminating hallucination as a complex reasoning task and propose a multi-agent debate approach to encourage divergent-thinking. Our approach can interpret why hallucination occurs and identify which image region caused the hallucinated descriptions.
    %\item Our approach can be applied to any MLLM in a black-box manner, offering a ready-to-use solution that requires no additional training to mitigate hallucinations.
%\end{itemize}

\section{Related Work}
\noindent\textbf{Hallucination in MLLM}
With the rapid advancement of MLLMs, hallucinations in these models have garnered increasing attention. Current research primarily focuses on the detection and elimination of hallucinations. \cite{li2023objecthallucination} treats hallucination detection as a binary classification issue, which limits its evaluation for open-ended responses. HalDetect \cite{gunjal2024detecting} identifies hallucinations by training a specialized classifier.

For hallucination mitigation, previous work generally either collects more high-quality visual instruction tuning data~\cite{sun2023, yu2024hallucidoctor} or leverages an extra correction model~\cite{gunjal2024detecting, wang2024vigc} to alleviate hallucinations. LLaVA-RLHF~\cite{sun2023} employs manually annotated data as reward signals to guide the model in generating less hallucinative responses. In~\cite{wang2024vigc}, the Visual Instruction Correction (VIC) model is introduced, adopting an iterative update mechanism to correct responses. Recently, ~\cite{yu2024hallucidoctor} categorized hallucinations into three types based on image content: object category, object attribute, and object relation, and proposed a fine-grained hallucination mitigation approach. Later, ~\cite{jiang2024haleval} introduced another type, termed \emph{event hallucination}, which involves inventing a fictional target and weaving an entire narrative around it.

\noindent\textbf{Multi-agent Debate}
Although LLMs have demonstrated remarkable performance in various tasks, they still struggle with complex reasoning tasks, driving research into the cognitive behaviors of LLMs. Initially, Chain-of-thought (CoT)~\cite{wei2023chainofthought} was proposed to enhance the reasoning ability of LLMs. Subsequently, approaches based on self-evaluation have gained increasing attention, such as Self-refine~\cite{madaan2023}, Tree-of-Thoughts~\cite{yao2023tree}, and Self-reflection~\cite{shinn2023}. These methods emphasize the process of introspection and examination of one's own thoughts. However, relying on a single perspective can introduce bias and instability in the results.
Inspired by the \emph{Collective Intelligence}~\cite{collectiveintelligence} and \emph{Role-playing}~\cite{li2023camel} strategies, the Multi-Agent Debate (MAD) framework~\cite{liang2024divergentthink} was introduced, wherein multiple agents articulate their arguments in a ``tit for tat" fashion. However, MAD has yet to be explored in the context of addressing the MLLM hallucination problem.

\section{Our Approach}
%Hallucination in MLLMs is often defined as the cross-modal inconsistency, \emph{i.e.,} a mismatch between the generated description and the input image. 
%Object is always the centre for the image description, thus some MLLM hallucination 
In this paper, we argue that hallucination in MLLMs stems from MLLMs' limitations in slow-thinking and divergent-thinking. Consequently, we propose an approach to mitigate MLLM hallucination by leveraging self-reflection and multi-agent debate strategies. In the following sections, we will first outline our multi-agent debate workflow and then present the specific prompt engineering method. Additionally, %we will describe how to seamlessly incorporate self-reflection into the debate procedure.
we will give a concrete example to illustrate the debate workflow.

\subsection{Workflow for Multi-agent Debate}
We follow a general multi-agent debate framework involving several MLLM-based agents as debaters. These debaters engage in a multi-round debate around one debate question. In practice, we adopt questions from the POPE benchmark as debate questions. These questions take the form ``Is there an \emph{object} in the image?" where the ``\emph{object}" is a specific object (\emph{e.g.,} ``clock") related to hallucination evaluation. This type of \emph{Yes}-or-\emph{No} question is ideal for debate. We argue that a multi-agent debate procedure can significantly reduce hallucination about this object and yield a more accurate answer.

Note that multi-agent debate is merely a framework for collaboration among different individuals, and we need to design a specific debate workflow for a particular cognitive task. For the task of hallucination mitigation, we propose to instantiate three MLLM-based agents: two debaters and one judge. Most of the debate occurs between the two debaters. If they cannot reach an agreement after several rounds of debate, the judge intervenes to make a final decision.

% \begin{table*}[!h]
% \begin{center}
% % \renewcommand\arraystretch{1.5}
% \caption{Prompt for Agents}
% \begin{tabular}{p{16cm}}
% \hline
% You are a debater. Hello and welcome to the visual question- answering competition, which will be conducted in a debate format. You can agree or disagree with others' viewpoints, as our objective is to find the correct answer. The criteria for a good answer are as follows:
% 1. Ensure that your output aligns with the content in the image. 2. Only output content that is visible in the image. 3. Distinguishing the nuances between synonyms is crucial.     
% \\ \hline 
% \end{tabular}
% \end{center}
% \end{table*}

\begin{figure}[tbp]
    \centering
    \includegraphics[width=1\linewidth]{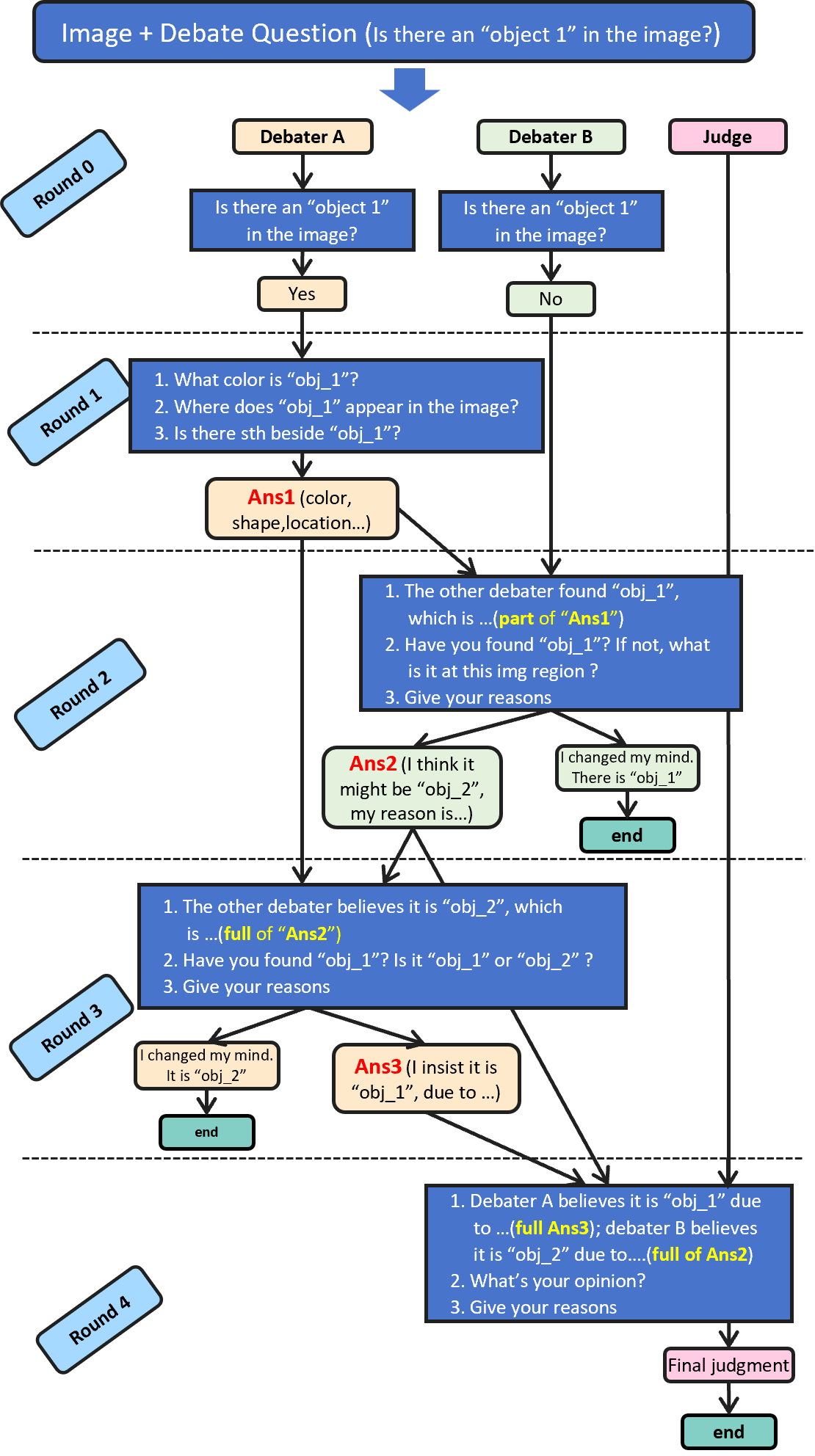} 
    \caption{Workflow of our multi-agent debate, wherein two debaters and one judge engage in a multi-round discussion addressing a hallucination-related debate question.
    }
    \label{fig2}
\vspace{-1.0em}
\end{figure}

\noindent\textbf{Self-reflection.} For each debater, we will ask a series of questions based on their answers to previous questions. This \emph{continuous inquiry} strategy is a commonly adopted self-reflection strategy. It encourages the agent to engage in slow-thinking and helps in obtaining detailed reasons. 

This strategy can be utilized at various stages of the debate.
As shown in Figure \ref{fig2}, in Round 1 we will ask debater A three types of questions if he believes the object-of-interest (\emph{e.g.,} $Obj_1$) is present in the input image. The first type pertains to the attributes of $Obj_1$, such as color, shape, and size. The second type concerns the location of $Obj_1$, indicating where the object appears in the image, such as the top-left or bottom-right of the image. The third type relates to the relations among objects, for example, what object is beside or on the left side of $Obj_1$. These \emph{continuous inquiry} will help to identify which image region causes the hallucination and why it occurs (\emph{e.g.,} due to mistakenly perceived color, shape, object relation, \emph{etc}).

\noindent\textbf{Multi-agent Debate.} 
Multi-agent debate aims to encourage the divergent-thinking through the collaboration of different individuals. The philosophy of the ``society of minds" posits that different individuals possess distinct advantages and suggests leveraging their collaboration to enhance overall performance.

In our approach, we adopt a role-playing scheme by assigning distinct personality to different debaters. Specifically, we designate Debater A as \emph{conservative}, focusing on objectively depicting the image content and describing objects they are highly confident about. Conversely, Debater B is assigned a more \emph{imaginative} personality, allowing him to utilize the surrounding context of the object to hypothesize about what this object might represent. This strategic differentiation of debaters’ personality enhances divergent-thinking and facilitates a more effective debate procedure.

Furthermore, cross-agent message propagation is also critical to the success of the debate. During the debate, we will propagate one debater's answer to the other. As shown in Fig.\ref{fig2}, in Round 2, we summarize debater A's answer and propagate them to hint at debater B. In Round 3, we summarize debater B's answer and propagate them back to debater A. 
%Moreover, if the two agents cannot reach an agreement, it is necessary to introduce a judge to make a decision by considering all answers from both agents. This message propagation encourages the debaters not only to compare the distinct answers but also to re-evaluate their previous responses.
Interestingly, our experimental results indicate that propagating \emph{all} information from one agent to the other is not optimal. Particularly in the early debate stages, it is necessary to share less information between debaters. For example, in Round 0, it is crucial to ask each debater the debate question independently. Otherwise, one debater's answer could significantly influence and mislead the other debater's response. Another example is for Round 2, if we propagate the \emph{full} answer of debater A to B, Debater B will be overwhelmingly convinced by A and quickly change their answer, thus losing the purpose of the debate. Therefore, we just select \emph{partial} information from A's answer and propagate them to B. Through experimental evaluation, we found that selecting only the object attribute (rather than the location information) is a good trade-off in practice. 

In contrast, at the later debate stage (\emph{e.g.,} Round 3), it is better for Debater A to obtain \emph{full} feedback from Debater B. Similarly, at the judge stage, it is necessary for a judge to make a decision by obtaining all answers from both agents. We will discuss this in detail in our experiments.

%Another example in Figure \ref{f-judge} shows the judgment stage (\emph{e.g.,} Round 4), where it is critical for the judge to make a good decision by utilizing all answers from Debater A and B.

\subsection{Prompt-engineering for Multi-agent Debate}
The prompts fed to MLLMs will significantly influence their behaviors. Since the debaters/agents in our approach are implemented using MLLMs, we need to carefully design distinct prompts for each debater at different stages/rounds of the debate.

To find the most effective prompts, we conduct extensive prompt engineering experiments. In our experiments, we use Gemini-Pro-Vision as our MLLM agent, though it is noteworthy that our prompt engineering methods can be applied to other MLLMs, such as GPT-4o. Here, we describe some key points for designing the prompts.

\begin{figure}[tpb]
 \centering
 \includegraphics[width=0.5\textwidth]{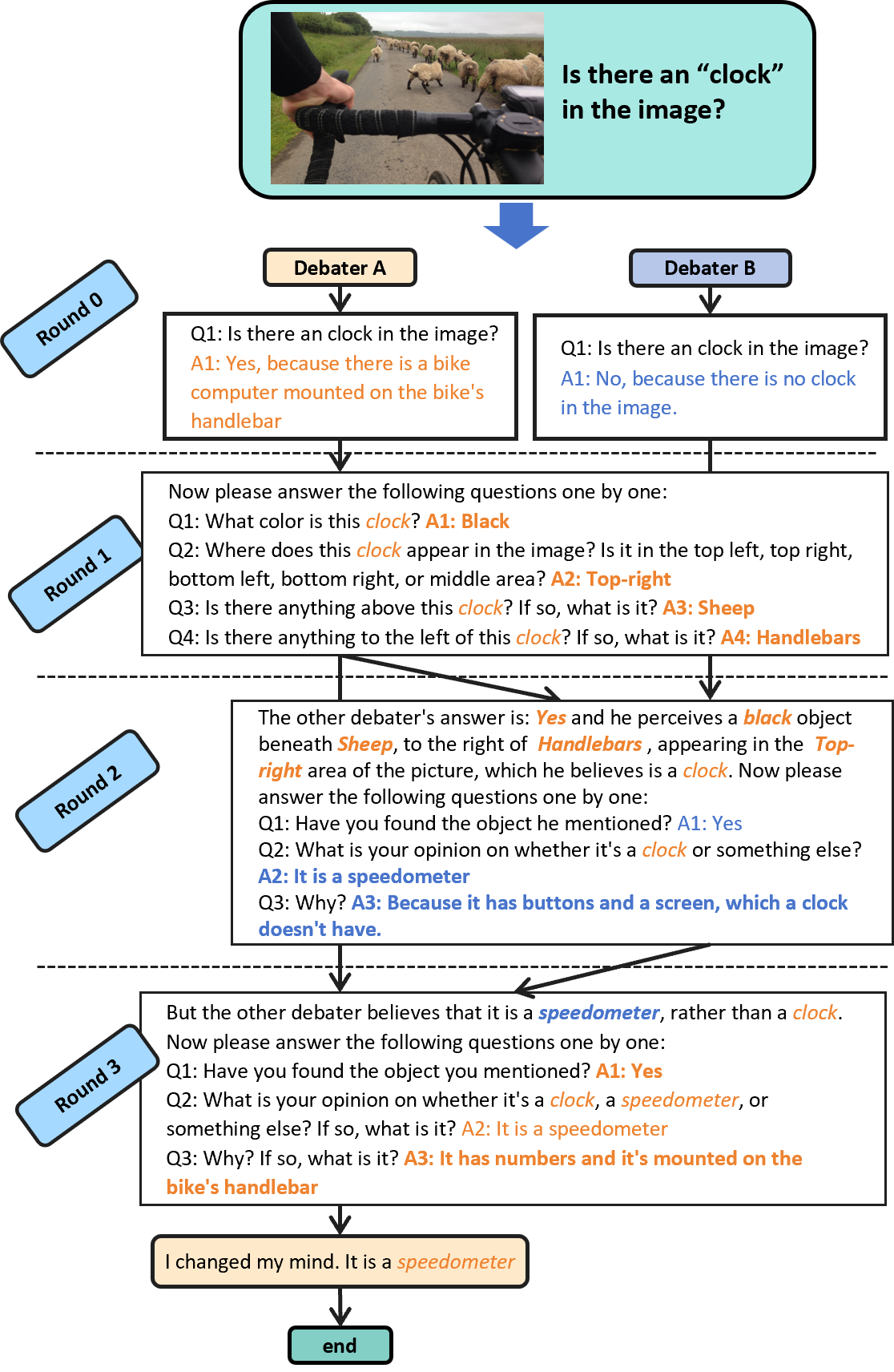}
 \caption{One example for our debate workflow.}
 \label{fig3}
\vspace{-1.0em} 
\end{figure}

\noindent\textbf{Pose Questions Strategically:} In Round 1, our objective is to gather detailed information about the object-of-interest from debater A, such as its precise location within the image. Recent advancements in MLLMs, including state-of-the-art models like Gemini-Pro-Vision or GPT-4o, have shown limitations in accurately pinpointing object locations. When directly asked, ``where does the object appear in the image?", MLLMs often provide incorrect responses. To address this challenge, we simplify the query by requesting coarse location information. Practically, we divide the image into five regions, \emph{i.e.,} center, top-left, top-right, bottom-left, and bottom-right, and inquire about the region where the object is located. This approach allows most MLLMs to provide accurate responses to these questions, as shown in Fig.\ref{fig3}.

\begin{figure*}[htbp]
\centering
\subfloat[a \emph{plate} is mislabeled as a \emph{bowl}]{
\includegraphics[width=0.172\linewidth]{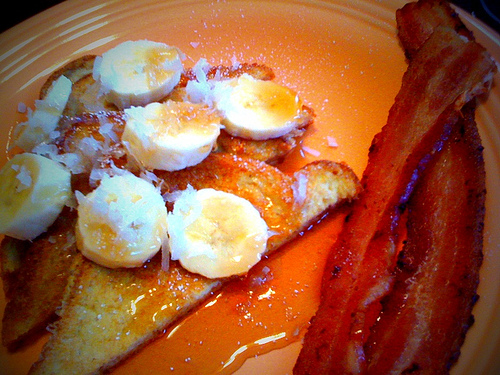}
}
\quad
\subfloat[a \emph{monitor} is mislabeled as a \emph{TV}]{
\includegraphics[width=0.172\linewidth]{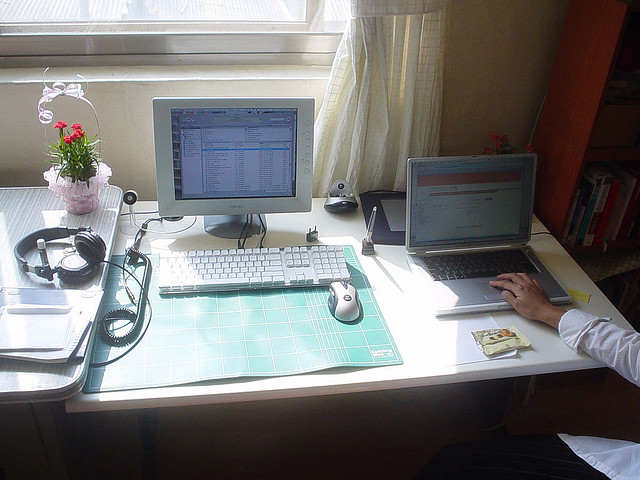}
}
\quad
\subfloat[confusion between ``car" and ``carriage"]{
\includegraphics[width=0.172\linewidth]{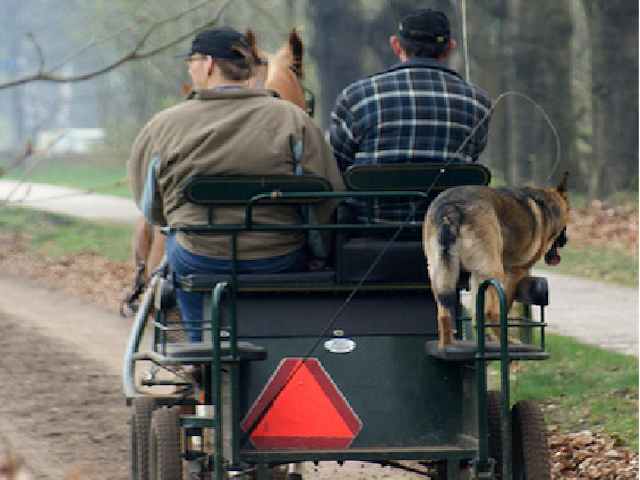}
}
\quad
\subfloat[hard to judge whether the table is a \emph{dining table} or not]{
\includegraphics[width=0.172\linewidth]{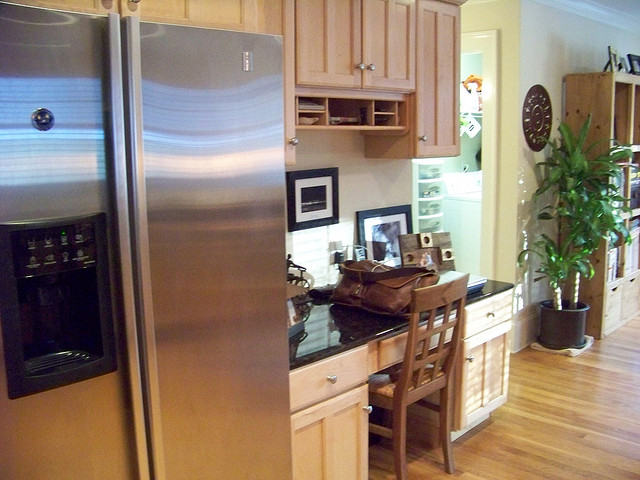}
}
\quad
\subfloat[hard to judge whether the cat is lying in a \emph{bed} or something else]{
\includegraphics[width=0.172\linewidth]{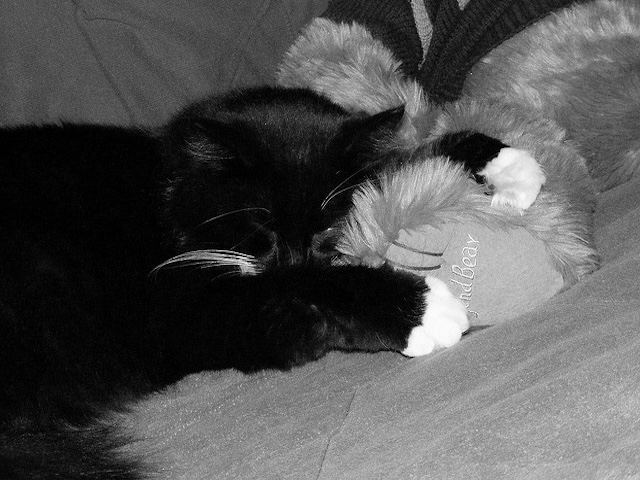}
}
%\quad
\caption{There are erroneous annotations and ambiguous samples in the original POPE benchmark.}\label{fig4}
\vspace{-1.0em}
\end{figure*}

\noindent\textbf{Imitation-based Prompting:} Inspired by imitation learning, we supply the debaters with a successful-and-relevant debate example to guide them in conducting a proper debate. Specifically, we collect several successful debate examples ahead and categorize them into representative scenarios such as \emph{sports}, \emph{kitchen}, \emph{office}, etc. When mitigating hallucination for a specific scenario, we select a matching example and use it as context to aid the debaters. Practically, we incorporate the selected example into the prompts and leverage MLLMs' in-context learning to utilize the provided example to enhance the quality of debate, as shown in Fig.~\ref{fig10}.
The underlying rationale for our imitation-based prompting method is that a good debate example can significantly aid debaters in performing their debate activity.

\subsection{One Example for Our Debate Workflow}

Fig.\ref{fig3} illustrates an instance of hallucination mitigation using our multi-agent debate approach. Without our debate process, agent A mistakenly identifies a ``speedometer" as a ``clock". However, through our debate procedure, agent A eventually observes that the object in the image possesses buttons and a screen, leading to the conclusion that it cannot be a clock. Furthermore, the object's position (\emph{i.e.,} on the bike's handlebar) further helps to identify it as a speedometer. This example vividly showcases the efficacy of our approach in interpreting the cause of hallucination and mitigating it.

\section{Interpreting Hallucination in MLLM}
Besides mitigating hallucination, our approach also offers the advantage of providing detailed interpretations for each hallucination case. This allows us to analyze many hallucination cases and gain insights into MLLM hallucination. Specifically, extensive analysis reveals that MLLM hallucination can be categorized into several types: 

(1) MLLMs often \emph{lack the capability} to accurately perceive image content. For example, as shown in Fig.\ref{fig1}(a), MLLM mistakenly identified a ``ski pole" as a ``fork" due to limited perceptual capability. 

(2) The object in the image looks like another object so that it becomes challenging for the MLLM to distinguish them. As shown in Fig.\ref{fig1}(b), MLLM mistakenly identified a ``plate" as a ``bowl" due to their \emph{visual similarity}.

(3) Two concepts are closely related and difficult to distinguish. As shown in Fig.\ref{fig1}(c), MLLM mistakenly identified a ``backpack" as a ``briefcase" due to their \emph{conceptual similarity}.

(4) MLLMs \emph{misunderstand} the questions' meaning. As shown in Fig.\ref{fig1}(d), MLLM misunderstands the fruit of ``orange" in the question as a color.

(5) MLLMs make \emph{unreasonable or excessive guesses/inferences}, hallucinating objects based on the perceived contextual elements. As shown in Fig.\ref{fig1}(e), MLLM mistakenly inferred the existence of a ``bed" upon perceiving a ``mattress". As shown in Fig.\ref{fig1}(f), MLLM mistakenly inferred the existence of a ``laptop" upon perceiving a ``keyboard".

With these insights, we can identify the limitations of current MLLMs and find promising directions for their improvement.

\section{Revised Hallucination Benchmark}
Various benchmarks and metrics have been proposed for evaluating hallucination. In this paper, we adopt the POPE benchmark~\cite{li2023evaluating} as it aligns well with our debate approach. However, even popular benchmarks have limitations. Through extensive analysis, we observed two limitations in the POPE benchmark. 

First, we identify a few incorrect annotations in the POPE. As shown in Fig.\ref{fig4}(a) and (b), a \emph{plate} is mislabeled as a \emph{bowl}, and a \emph{computer monitor} is mislabeled as a \emph{TV}. Additionally, note that confusing labels will also lead to misannotations. For example, annotators often make mistakes when distinguishing between ``carriage" and ``car", as shown in Fig.\ref{fig4}(c). Therefore, we have amended the POPE benchmark by correcting these erroneous annotations, resulting in the revised benchmark, called \emph{POPE-R}. 

Secondly, we have observed that some cases are too difficult, \emph{even for humans}, to judge whether hallucination occurs. As depicted in Fig.\ref{fig4}(d), it is challenging for humans to judge whether the table is a \emph{dining table} or not, since there is no food but a bag on the table. Therefore, we filter out these \emph{ambiguous} cases in our POPE-R benchmark.

Furthermore, we found that the current criteria for hallucination evaluation are somewhat coarse, failing to distinguish between creativity and hallucination.

\subsection{Creativity vs. Hallucination}
Recently, some research has proposed distinguishing creativity from hallucination in LLMs. For instance, ~\cite{jiang2024} asked, ``\emph{Is hallucination in LLMs always harmful, or does creativity hide in hallucinations?}" Furthermore, ~\cite{zhao2024assessing} claims that ``\emph{the creativity they exhibit is a key reason they are considered powerful}," and proposes a method to harness hallucination for creativity.

\begin{figure}[tbp]
\centering
\subfloat[We can infer the existence of a ``TV" from observing a girl holding a ``Wii" remote.]{
\includegraphics[width=0.44\linewidth]{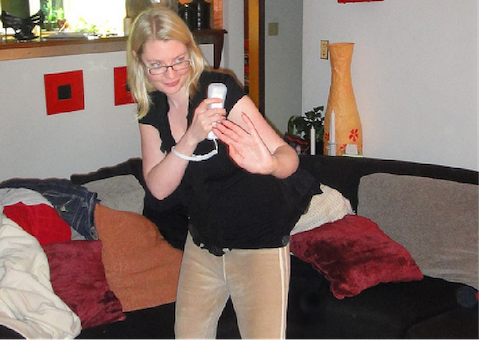}
}
\quad
\subfloat[We can infer the presence of a ``spoon" from the observed set of ``tableware".]{
\includegraphics[width=0.44\linewidth]{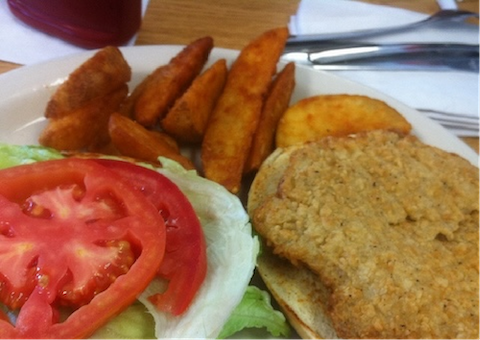}
}
%\quad
\caption{Creativity vs. Hallucination.}\label{fig5}
\vspace{-1.0em}
\end{figure}

In the context of MLLMs, their creativity often manifests through \emph{association} and \emph{imagination}. Notably, we have observed cases where an object is not directly visible in the given image but can be inferred from other visible objects. As shown in Fig.\ref{fig5}(a), the question is ``Is there a \emph{TV} in the image?" Although a TV is not directly visible, we can see a girl playing a Wii game (a type of TV game), leading us to infer the presence of a TV. Fig.\ref{fig5}(b) is another example illustrating that MLLM can creatively infer the existence of a ``spoon" upon perceiving a set of ``tableware". Hence, we argue that this phenomenon reflects the MLLM's advantage in creativity rather than its  drawback in hallucination. 
Therefore, we propose to distinguish MLLMs' creativity from their hallucinations.

Furthermore, we propose a method to evaluate MLLMs' creativity capabilities. Specifically, we manually identify these \emph{associative} and \emph{imaginative} cases, resulting in a creativity benchmark termed the \emph{POPE-C}. Consequently, we can leverage POPE-C to evaluate the creativity capacities of MLLMs.

\section{Evaluation}
\subsection{Implementation}
\noindent\textbf{Multimodal LLMs.} 
In our approach, we utilize MLLMs as debaters/agents. Recently, numerous MLLMs have been proposed, including 
open-sourced models such as LLaVA-1.5~\cite{liu2024improved} and MiniGPT-4~\cite{zhu2023minigpt}, as well as close-sourced models like Gemini-Pro-Vision~\cite{team2023gemini} and GPT-4o. 
However, we found that most open-sourced models are incapable of engaging in debate activities.
Some models fail to understand our debate prompts and do not know how to properly engage in the debate. Other models are too weak, easily misled by other debaters, resulting in premature agreement, thus failing to sustain a multi-round debate.

After extensive exploration, we discovered that two models are qualified to act as debaters: Gemini-Pro-Vision and GPT-4o. Thus, we utilize them as debaters in this paper. 
Note that we have found that the more powerful an MLLM is, the better it engages in debate activities. Thus, as MLLMs continue to advance, our approach is expected to exhibit even greater advantages.
%Unless otherwise stated, we utilize Gemini-Pro-Vision as the default MLLM for all agents and set the temperature to $0$ to ensure reproducibility. Their accessibility and user-friendliness through APIs allow us to directly invoke and interact with the models during our research, greatly simplifying the process.

\subsection{Evaluation Benchmark and Metrics.}
\noindent\textbf{POPE.} 
The benchmark for evaluating hallucination can be categorized into two types: generative and discriminative benchmark. Discriminative benchmark often transforms the evaluation of hallucination into a binary classification task by prompting MLLMs with simple \emph{Yes}-or-\emph{No} questions about the probing objects. We adopt this kind of benchmark in our evaluation, as we can directly treat the \emph{Yes}-or-\emph{No} questions as debate questions.

One representative discriminative benchmark is the Polling-based Object Probing Evaluation (POPE)~\cite{li2023evaluating}, which uses a question format like ``Is there an \emph{object} in the image?" to determine whether a model has hallucination for the probing \emph{object}. The comprehensive POPE test consists of three sampling settings: (1) the ``random" setting, which determines the probing objects by randomly sampling all objects present in the entire dataset; (2) the ``popular" setting, which selects the most frequently occurring objects in the dataset as the probing objects; (3) the ``adversarial" setting, which selects the objects highly relevant to the given image as the probing objects.

The POPE benchmark compiles data from three diverse sources: MSCOCO~\cite{lin2014microsoft}, A-OKVQA~\cite{schwenk2022okvqa}, and GQA~\cite{hudson2019gqa}. The proportion of queries probing existent and non-existent objects is evenly balanced (\emph{i.e.}, $50\%$ vs. $50\%$). The evaluation is based on five critical metrics: \emph{Accuracy}, \emph{Precision}, \emph{Recall}, \emph{F1 score}, and the \emph{Yes-ratio}.

\noindent\textbf{POPE-R.} 
We have found that POPE benchmark contains \emph{incorrect} and \emph{ambiguous} annotations, which can impact the quality of hallucination evaluation. Thus, we propose a revised benchmark, termed \emph{POPE-R}, which addresses these issues by correcting misannotations and filtering out ambiguous cases. We will conduct evaluations on both the original POPE and our revised \emph{POPE-R} benchmarks.

\noindent\textbf{POPE-C.} 
In this paper, we propose to distinguish creativity from hallucination and further evaluate MLLMs' creativity capabilities. The evaluation is conducted using our \emph{POPE-C} benchmark. 

Specifically, we manually identify the cases where the probing object is not directly visible in the given image but can be inferred from other visible objects. We pick out these cases to build the \emph{POPE-C} benchmark. By inquiring whether an MLLM can infer the existence of these objects, we can assess the model's associative and imaginative capabilities.
The more cases an MLLM can infer, the greater its advantage in creativity. Thus, we use the \emph{Creativity-ratio} as a metric to evaluate the creativity capability.

\subsection{Main Results}
Our approach integrates self-reflection and multi-agent debate strategies for hallucination mitigation. To assess their individual contributions, we propose another variant, termed the Self-Reflection-Only (SRO) approach, which removes the multi-agent debate strategy and retains only the self-reflection strategy.

\begin{figure}[t]
    \centering
    \includegraphics[width=0.7\linewidth]{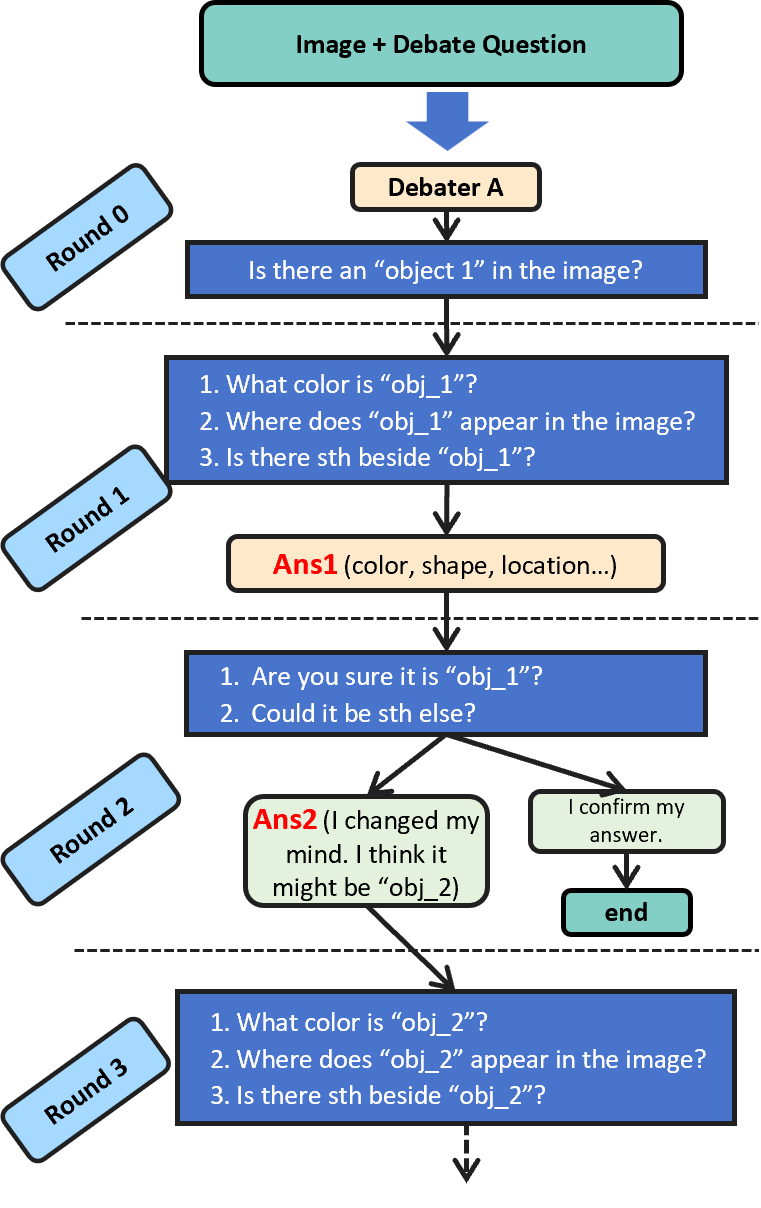} 
    \caption{Workflow of Self-Reflection-Only (SRO) approach.}
    \label{fig6}
\vspace{-1.0em}    
\end{figure}

The specific workflow of SRO is shown in Fig.\ref{fig6}. Here, only one agent is involved in hallucination mitigation. While it still entails several rounds of discussion, this discussion occurs within the agent itself. After answering a question, the agent poses additional detailed questions to itself and re-evaluates its previous response. This self-reflection process encourages the agent to engage in slow-thinking, thereby mitigating hallucinations.

In this section, we compare three approaches: the \emph{Baseline} approach, which directly leverages an MLLM with neither self-reflection nor multi-agent debate strategies; the \emph{SRO} approach, which only employs the self-reflection strategy; and \emph{Our} approach, which incorporates both self-reflection and multi-agent debate strategies.
We compare the three approaches on both POPE and POPE-R datasets. Following~\cite{li2023evaluating}, we adopt three settings: Random, Popular and Adversarial settings.

\begin{table*}[htbp]
\begin{center}
\renewcommand\arraystretch{1.2}
\caption{Comparison on hallucination mitigation among the \emph{Baseline}, the \emph{SRO} and \emph{Our approach} under three evaluation settings of POPE dataset (the MSCOCO validation set). The metric of \emph{Yes-ratio} denotes the proportion of answering ``Yes" to the given question. The best results in each block are highlighted in bold.}
\resizebox{0.8\textwidth}{!}{
\begin{tabular}{c c c c c c | c | c}
\toprule[1.5pt]%设置线的宽度
\textbf{Models}        & \textbf{Settings}                & \textbf{Methods} & Accuracy & Precision & Recall & F1 Score & Yes-ratio \\ 
\midrule
\multirow{9}{*}{Gemini-Pro-Vision} & \multirow{3}{*}{Random}      & Baseline  &    85.20      &     94.82      &     74.47   &    83.42      &   39.27      \\      
                        &                             & SRO   &     86.00     &    92.86       &     78.00   &      84.78    &    \textbf{42.00}     \\ 
                        &                             & Ours   &     \textbf{87.53}     &    \textbf{95.18}       &     \textbf{79.07}   &      \textbf{86.38}    &    41.53     \\ \cmidrule{2-8}
                        & \multirow{3}{*}{Popular}     &    Baseline            &  83.47        &    \textbf{90.68}       &  74.60      &    81.86      &    41.13     \\
                        &                              &     SRO           &   84.10       &        88.14   &   78.80     &   83.21       &    44.70     \\
                        &                              &     Ours           &  \textbf{85.30}        &     87.63      &   \textbf{82.20}     &    \textbf{84.83}      &      \textbf{46.90}   \\ \cmidrule{2-8}
                        & \multirow{3}{*}{Adversarial} &     Baseline           &    82.17      &    \textbf{87.61}       &    74.93    &    80.78      &  42.77       \\
                        &                              &     SRO            &    82.93      &      85.19     &  79.73      &    82.37      &   46.80      \\
                        &                              &     Ours           &  \textbf{84.43}        &     86.66      &    \textbf{81.40}    &    \textbf{83.95}      &   \textbf{46.97}      \\ \cmidrule{1-8}
\multirow{9}{*}{GPT-4o} & \multirow{3}{*}{Random}      & Baseline  &    85.77      &   92.14        &   78.20     &   84.50       &    \textbf{42.43}     \\      
                        &                             & SRO   &    86.83      &       93.68    &    \textbf{79.00}    &  85.71        &   42.17      \\ 
                        &                             & Ours   &  \textbf{87.83}        &    \textbf{98.88}       &    76.53    &    \textbf{86.28}      &    38.70     \\ \cmidrule{2-8}
                        & \multirow{3}{*}{Popular}     &    Baseline            &  84.20        &       88.80    &  \textbf{78.27}      &   83.20       &   \textbf{44.07}      \\
                        &                              &     SRO           &  85.27        &   91.07        &    78.20    &     84.15    &   42.93      \\
                        &                              &     Ours           &    \textbf{86.73}      &  \textbf{97.91}         &    75.07    &   \textbf{84.98}       &  38.33       \\ \cmidrule{2-8}
                        & \multirow{3}{*}{Adversarial} &     Baseline           &    83.00      &       85.97    &     \textbf{78.87}   &       82.27   &    \textbf{45.87}     \\
                        &                              &     SRO            &     84.10     &   89.99        &  76.73      &  82.84        &   42.63      \\
                        &                              &     Ours           &   \textbf{85.87}       &    \textbf{97.61}       &   73.53     &  \textbf{83.88}        &    37.67     \\ 
\bottomrule[1.5pt]%设置线的宽度
\end{tabular}}
\label{tab:t1} % 为表格创建标签
\end{center}
\vspace{-1.0em}
\end{table*}

\noindent\textbf{Results on POPE.} 
Table \ref{tab:t1} shows the comparison results on POPE under the three settings. We can see that our approach comprehensively outperforms the baseline and SRO. For example, our approach improves the accuracy from $85.2\%$ to $87.53\%$ for the random setting, while the SRO achieves an accuracy of $86\%$. Particularly, our approach demonstrates greater advantages than the SRO in the adversarial setting ($84.43\%$ vs. $82.93\%$). 
More importantly, even if our approach cannot mitigate some hallucination cases, it still can identify which image region leads to the hallucination and interpret why such hallucinations occur. %Through the analysis of bad cases, some incorrect predictions even possess convincing justifications, such as more comprehensive and nuanced answers and creative interpretations of the image that differ from hallucinations. 
% And more examples are also given in the Appendix xxx.

\noindent\textbf{Results on POPE-R} 
Table \ref{tab:t2} presents the results of baseline, SRO and our approach on POPE-R under the random setting. After correcting numerous erroneous labels in the original POPE, it is apparent that the three methods have achieved significant improvements compared to their previous performances. Our approach, in particular, stands out with a 3.07\% increase in ``Accuracy" and a 3.96\% increase in ``F1 Score" over the baseline, exceeding even the gains achieved on the original POPE. This notable enhancement not only validates the effectiveness of our approach but also emphasizes the importance of addressing the errors and inconsistencies present in the original dataset.

% \noindent\textbf{Interpreting MLLM Hallucination:}
% 这个部分写什么

%\begin{figure}[!h]
    %\centering
    %\includegraphics[width=1\linewidth]{pic/multi > single/cell phone wii remote.png}
    %\caption{  The comparison illustration between the individual model and Multi-Agent Debate.}
    %\label{wii remote-case}
%\end{figure}

\begin{table*}[htbp]
\begin{center}
\renewcommand\arraystretch{1.2}
\caption{Comparison on hallucination mitigation among the \emph{Baseline}, the \emph{SRO} and \emph{Our approach} with our revised POPE-R dataset.}
\resizebox{0.8\textwidth}{!}{
\begin{tabular}{c c c c c c | c | c}
\toprule[1.5pt]%设置线的宽度
\textbf{Models}        & \textbf{Settings}                & \textbf{Methods} & Accuracy & Precision & Recall & F1 Score & Yes-ratio \\ 
\midrule
\multirow{3}{*}{Gemini-Pro-Vision} & \multirow{3}{*}{Random}      & Baseline  &    91.02      &    92.41       &  85.68      &       88.92   &    39.27     \\
                         &                             & SRO   & 91.06         &     89.90      &   88.69     &    89.29      &    \textbf{42.00}     \\
                        &                             & Ours   & \textbf{94.09}         &     \textbf{94.07}      &   \textbf{91.71}     &    \textbf{92.88}      &    41.53    \\ \cmidrule{1-8}
\multirow{3}{*}{GPT-4o} & \multirow{3}{*}{Random}      & Baseline  &     91.20     &     89.60      &   89.45    &    89.52      &    \textbf{42.43}     \\
                         &                             & SRO   &     91.97     &    90.73       &  \textbf{90.12}      &     90.42     &  42.17       \\
                        &                             & Ours   & \textbf{94.54}         &     \textbf{98.33}      &    88.53    &    \textbf{93.17}      &   38.70     \\
\bottomrule[1.5pt]%设置线的宽度
\end{tabular}}
\label{tab:t2} % 为表格创建标签
\end{center}
\vspace{-1.0em}
\end{table*}

\begin{figure}[!h] 
    \centering
    \includegraphics[width=1\linewidth]{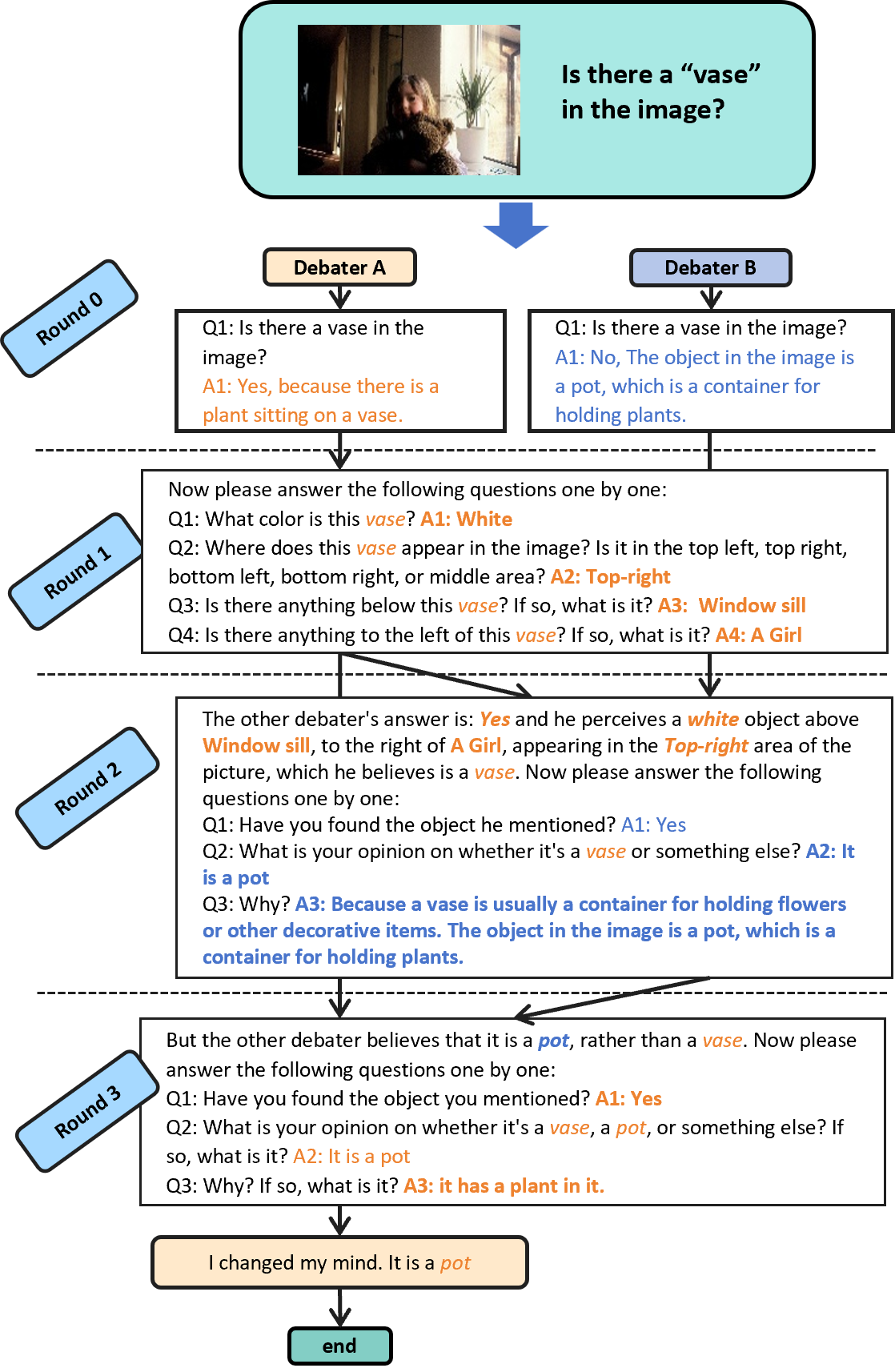}
    \caption{Our debate approach enables identifying the object in the image as a \emph{pot} rather than a \emph{vase}, along with a detailed explanation.}
    \label{fig7}
\end{figure}

\noindent\textbf{More Examples for Our Debate.} 
Fig.\ref{fig3} has shown an example of our debate workflow, and we will show more examples in this section. Fig.\ref{fig7} is another example, where the debate question is ``Is there a \emph{vase} in the image?" Initially, Debater A mistakenly answers ``Yes.". Through several rounds of debate, Debater A recognizes his mistake and revises his answer to ``No." This example clearly illustrates that the cause of A's hallucination is due to the \emph{conceptual} confusion between a \emph{pot} and a \emph{vase}. Furthermore, with our debate approach, Debater A is persuaded by Debater B's reason: ``A vase is usually a container for holding \emph{flowers}, while the object in the image is a pot, which is a container for holding \emph{plants}."

\begin{figure}[!h] 
    \centering
    \includegraphics[width=1\linewidth]{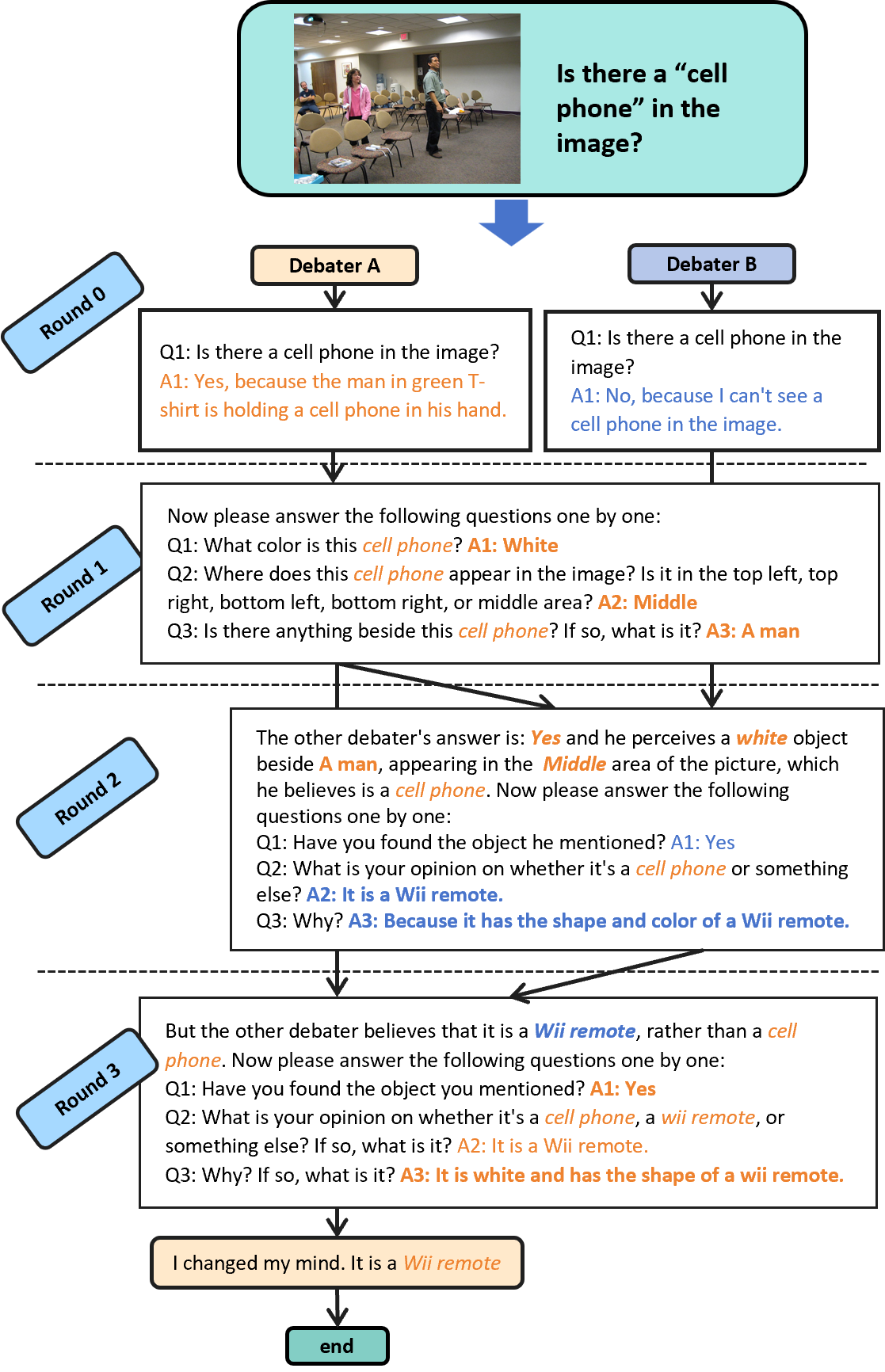}
    \caption{Our debate approach enables identifying the object in the image as a \emph{Wii remote} rather than a \emph{cell phone}.}
    \label{fig8}
\end{figure}

Figure \ref{fig8} illustrates one more example. Initially, Debater A incorrectly believed there was a cell phone in the image. Fortunately, with our continuous inquiry, the debate focus shifted from the entire image to a specific region, narrowing the original question to whether the object in the man's hand was a cell phone or a Wii remote. Consequently, it became easier to identify the object as a Wii remote rather than a cell phone based on its color and shape features. Clearly, our Multi-Agent Debate can narrow the scope of the debate question and zoom in on a specific image region, achieving significantly better results.

\begin{table}[htbp]
\begin{center}
\renewcommand\arraystretch{1.0}
\caption{Comparison on model's \emph{Creativity} capability among the \emph{Baseline}, the \emph{SRO} and \emph{Our approach} using our POPE-C benchmark. The metric of \emph{Creativity-ratio} denotes the proportion of creativity answer to the given question.}
\resizebox{0.42\textwidth}{!}{
\begin{tabular}{c c | c }
\toprule[1.5pt]%设置线的宽度
\textbf{Models}    & \textbf{Methods} & Creativity-ratio \\ 
\midrule
\multirow{3}{*}{Gemini-Pro-Vision}  & Baseline  &    46.15           \\
                         & SRO   &      53.85         \\
                         & Ours   &  \textbf{69.23}        \\ \cmidrule{1-3}
\multirow{3}{*}{GPT-4o}  & Baseline  &  23.08             \\
                          & SRO   &    30.77            \\
                        & Ours   &    \textbf{38.61}      \\
\bottomrule[1.5pt]%设置线的宽度
\end{tabular}}
\label{tab:t3} % 为表格创建标签
\end{center}
\vspace{-2.0em}
\end{table}

\subsection{Creativity}
% We illustrate an example demonstrating how our approach enhances the creative capabilities of MLLMs. As shown in Fig.\ref{remote tv-case}, based on the perceived \emph{Wii Remote} in the image, the debate procedure encourages a creative association between the \emph{Wii Remote} and a \emph{TV}, leading to the inference of a TV's presence.
In addition to mitigating hallucination in MLLMs, we propose evaluating their creativity capabilities, \emph{i.e.,} their ability to generate novel, meaningful, and logical ideas. It is important to note that the debate workflow can influence a model's creativity. In this section, we quantitatively compare different debate strategies to assess their impact on a model's creativity.

We conduct the evaluation using the proposed POPE-C benchmark. From Table \ref{tab:t3}, we can see that the baseline method (\emph{i.e.,} the original MLLMs without any debate strategy) has a low \emph{Creativity-ratio} score, indicating that their creativity capability is limited. 
In contrast, both SRO and our multi-agent debate strategies could significantly enhance the model's creativity capability. Moreover, our strategy outperforms the SRO strategy remarkably, improving the \emph{Creativity-ratio} score from 53.85\% to 69.23\%.

\subsection{Ablation and Discussions}
Our multi-agent debate workflow incorporates several techniques, including message propagation among debaters, imitation-based prompting, and role-playing. In this section, we will assess their contributions.

\noindent\textbf{Message Propagation among Debaters.}
Message propagation is crucial to our debate approach. We have found that propagating all available information from one agent to another is not optimal, particularly during the early stages of the debate.

\begin{figure*}[ht]
    \centering
    \includegraphics[width=0.8\linewidth]{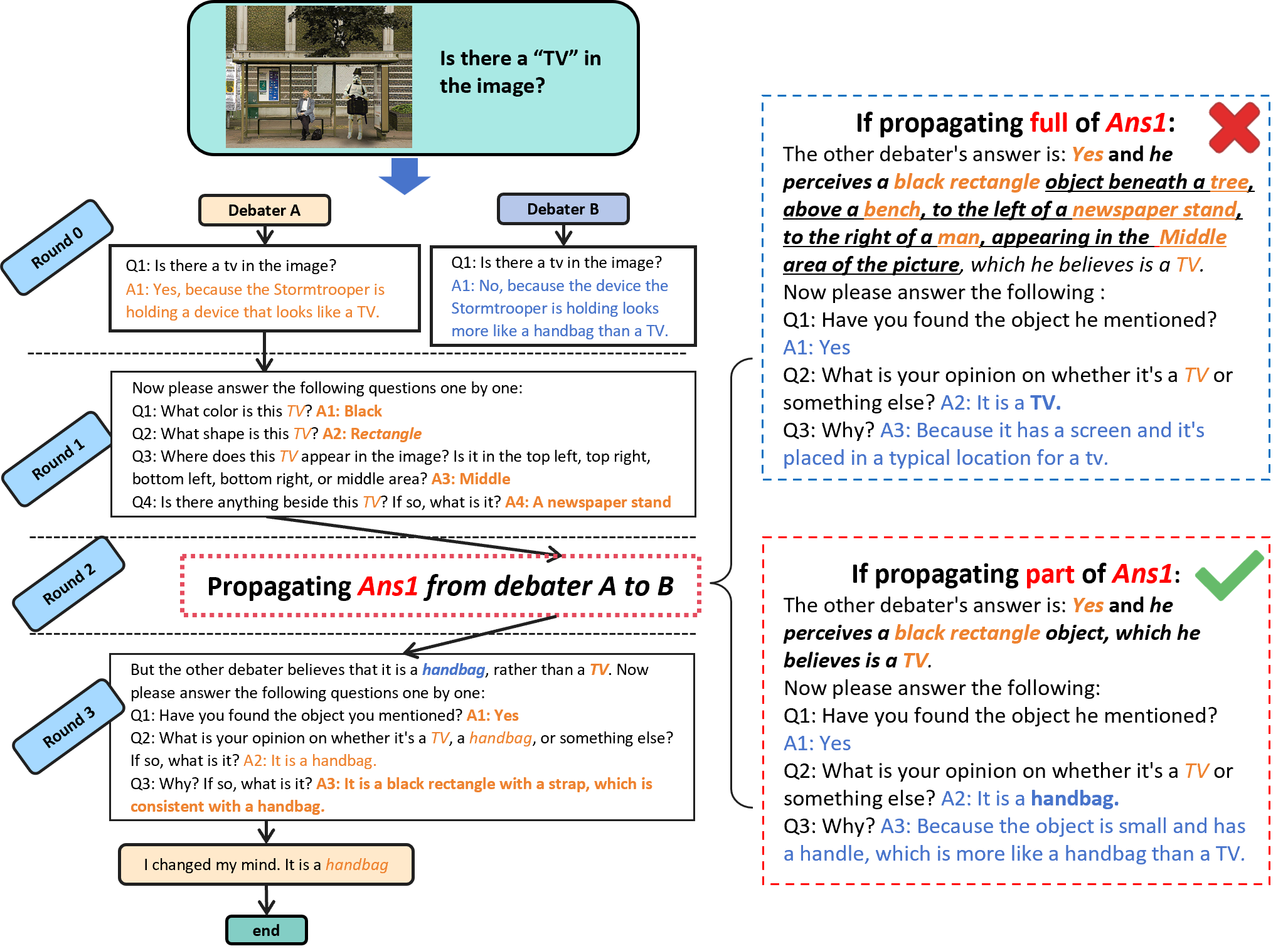} 
    \caption{If only a partial message of Agent A's \emph{Ans1} is propagated to Agent B in Round 1, Agent B will not be misled by Agent A and will stick to its original judgment.}
    \label{fig9}
\end{figure*}

As shown in Fig~\ref{fig9}, if we propagate only partial information (\emph{e.g.,} the object's color and shape) from Agent A to Agent B, Agent B is able to refute Agent A's answer and argue that the Stormtrooper is holding a handbag rather than a TV. In contrast, when propagating all message from Agent A to B, including a substantial amount of erroneous location information, Agent B is easily misled by Agent A and rapidly changes its answer, concluding that the Stormtrooper is holding a TV, thus undermining the effectiveness of the debate.

We argue that in the early stages of the debate, each debater tends to generate erroneous information. Thus, it is necessary to filter out unreliable messages to prevent misleading other debaters.

\noindent\textbf{Imitation-based Prompting.}
We propose to exploit imitation learning mechanism in our approach, which supplies a successful-and-relevant debate example to assist debaters in performing their debate task. 

Specifically, we first collect successful debate examples covering representative scenarios such as \emph{sports}, \emph{kitchen}, \emph{office}, etc. When mitigating hallucinations for a given image, we select a relevant example from this collection and use it to guide the debate. This is achieved by summarizing the selected example and incorporating its condensed discussion into the debaters' prompts, leveraging the MLLMs' strong in-context learning capability.

\begin{figure*}[htbp]
    \centering
    \includegraphics[width=0.8\linewidth]{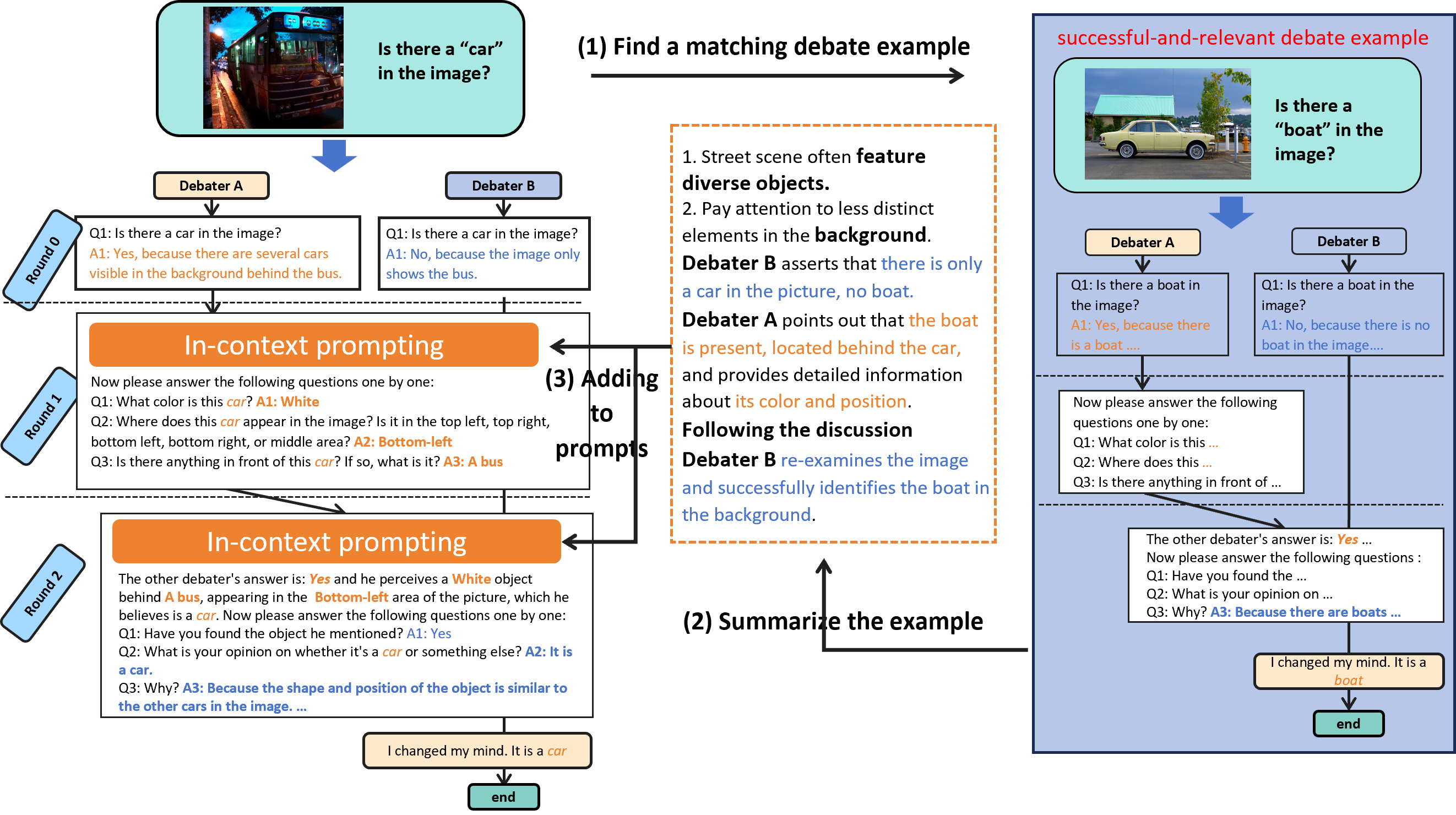} 
    \caption{On the right side is a successful-and-relevant debate example with respect to the left debate. We can incorporate this selected example into the debaters' prompts to assist them in performing their debate tasks.}
    \label{fig10}
\end{figure*}

As illustrated in Fig~\ref{fig10}, without the successful-and-relevant example, the debaters mistakenly focused solely on the primary subject, the \emph{bus}, in the image, overlooking the \emph{car} in the background. In contrast, with an example for the similar \emph{street} scenario, the debaters learned that such contexts often feature diverse objects, requiring careful observation, especially of less distinct elements in the background. Additionally, the debater grasped the idea that vehicles commonly appear in street scenes, prompting a more thorough reevaluation. As a result, the debater successfully identified the \emph{car} in the blurred background.

\noindent\textbf{Role-playing Strategy.}
In our approach, we assigned distinct personalities to different debaters. Specifically, Debater A is designated as \emph{conservative}, focusing on accurately understanding the image content and describing objects they are highly confident about. In contrast, Debater B is given a more \emph{imaginative} personality, allowing them to infer potential objects based on visible items or the depicted environment. This role-playing strategy enhances divergent thinking, intensifies discussions, and fosters creativity.

\begin{figure*}[htbp]
    \centering
    \includegraphics[width=0.8\linewidth]{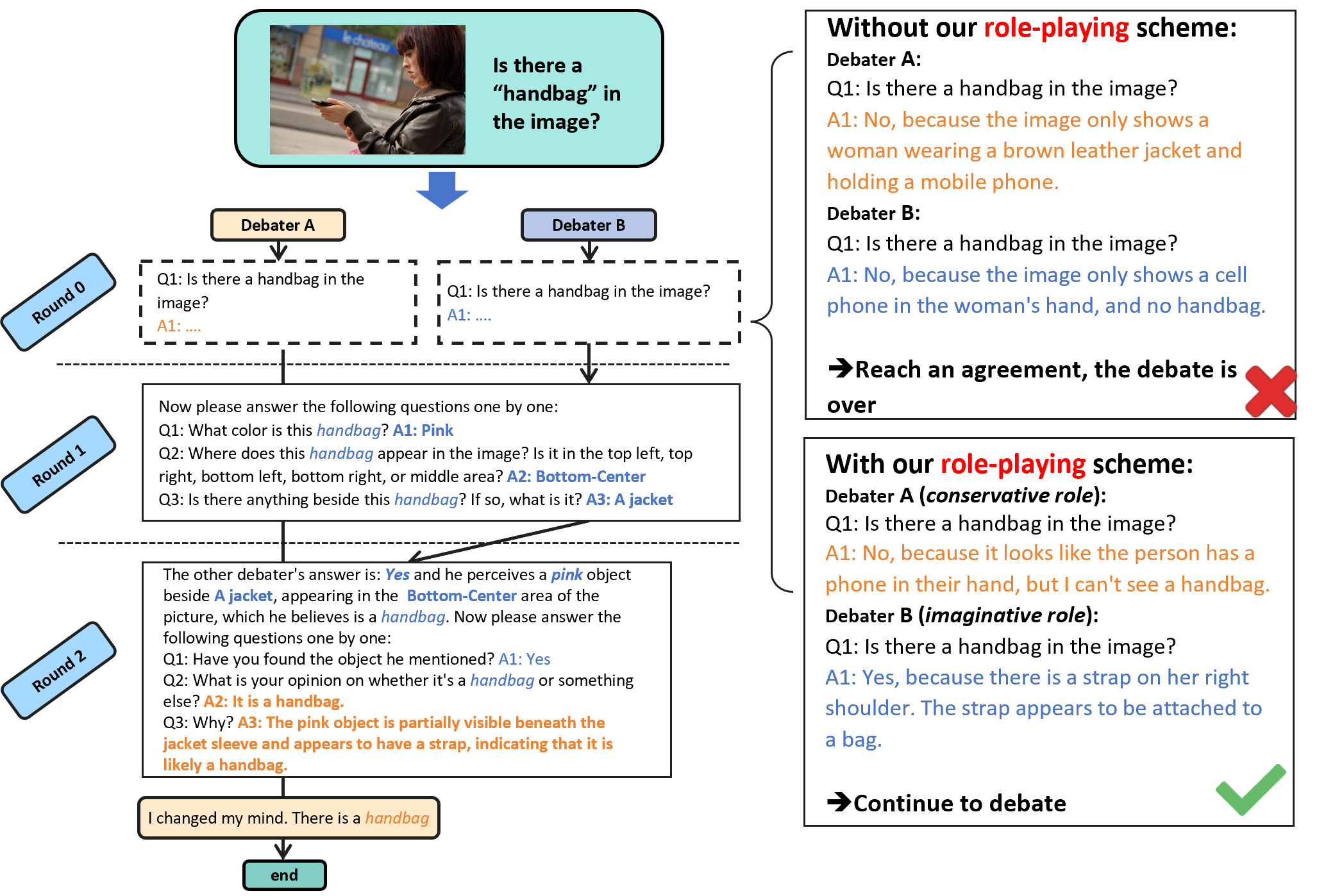} 
    \caption{Our role-playing strategy makes Agent B notice a strap on the person's right shoulder and deduce that the strap seems to be connected to a bag, leading to the inference of the existence of a \emph{handbag}.
    }
    \label{fig11}
\vspace{-1.0em}    
\end{figure*}

As shown in Fig~\ref{fig11}, without the role-playing strategy, both Agents A and B would overlook the pink handbag, which is only partially visible at the bottom of the picture, and incorrectly state that there is no \emph{handbag} present. In contrast, with our role-playing strategy, Agent B notices a strap on the person's right shoulder and deduces that it appears to be attached to a bag, leading to the inference of the \emph{handbag}'s existence. This triggers a further discussion that encourages Agent A to observe the picture more closely, ultimately changing A's perspective and acknowledging the presence of the \emph{handbag} in the image. We consider such discussions crucial to sustaining a multi-round debate.

% \subsection{Creativity}
% 这个部分写什么

%\begin{figure}[!h]
%    \centering
%    \includegraphics[width=1\linewidth]{pic/judge.png} 
%    \caption{For the judge to make an accurate decision, it is necessary to propagate all messages from Debater A and Debater B to the judge.}
%    \label{f-judge}
%\end{figure}

\section{Conclusion}
This paper delves into the causes of hallucinations in MLLMs, positing that hallucinations arise from the MLLMs' limitations in slow-thinking and divergent-thinking. To address this, we propose the adoption of self-reflection and multi-agent debate strategies for mitigating MLLM hallucinations. our approach can interpret why hallucination occurs and identify which image region caused the hallucinated descriptions. It can be applied to any MLLM in a black-box manner, providing a ready-to-use solution that requires no additional training. Furthermore, we propose distinguishing creativity from hallucination in the context of MLLMs and demonstrate how to evaluate MLLMs' creativity. Extensive experiments across various benchmarks and MLLMs confirm the effectiveness of our approach in reducing hallucinations.

%%%%%%%%% REFERENCES
{\small
\bibliographystyle{ieee_fullname}
\bibliography{egbib}

\begin{thebibliography}{10}\itemsep=-1pt

\bibitem{bai2024}
Zechen Bai, Pichao Wang, Tianjun Xiao, Tong He, Zongbo Han, Zheng Zhang, and Mike~Zheng Shou.
\newblock Hallucination of multimodal large language models: A survey, 2024.

\bibitem{chan2023chateval}
Chi-Min Chan, Weize Chen, Yusheng Su, Jianxuan Yu, Wei Xue, Shanghang Zhang, Jie Fu, and Zhiyuan Liu.
\newblock Chateval: Towards better llm-based evaluators through multi-agent debate, 2023.

\bibitem{chen2024slow}
Guangyao Chen, Siwei Dong, Yu Shu, Ge Zhang, Jaward Sesay, Börje~F. Karlsson, Jie Fu, and Yemin Shi.
\newblock Autoagents: A framework for automatic agent generation, 2024.

\bibitem{chu2024naviga}
Zheng Chu, Jingchang Chen, Qianglong Chen, Weijiang Yu, Tao He, Haotian Wang, Weihua Peng, Ming Liu, Bing Qin, and Ting Liu.
\newblock Navigate through enigmatic labyrinth a survey of chain of thought reasoning: Advances, frontiers and future, 2024.

\bibitem{dai2023instructb}
Wenliang Dai, Junnan Li, Dongxu Li, Anthony Meng~Huat Tiong, Junqi Zhao, Weisheng Wang, Boyang Li, Pascale Fung, and Steven Hoi.
\newblock Instructblip: Towards general-purpose vision-language models with instruction tuning, 2023.

\bibitem{gunjal2024detecting}
Anisha Gunjal, Jihan Yin, and Erhan Bas.
\newblock Detecting and preventing hallucinations in large vision language models, 2024.

\bibitem{hendrycks2021}
Dan Hendrycks, Collin Burns, Steven Basart, Andy Zou, Mantas Mazeika, Dawn Song, and Jacob Steinhardt.
\newblock Measuring massive multitask language understanding, 2021.

\bibitem{hudson2019gqa}
Drew~A Hudson and Christopher~D Manning.
\newblock Gqa: A new dataset for real-world visual reasoning and compositional question answering.
\newblock In {\em Proceedings of the IEEE/CVF conference on computer vision and pattern recognition}, pages 6700--6709, 2019.

\bibitem{jiang2024haleval}
Chaoya Jiang, Wei Ye, Mengfan Dong, Hongrui Jia, Haiyang Xu, Ming Yan, Ji Zhang, and Shikun Zhang.
\newblock Hal-eval: A universal and fine-grained hallucination evaluation framework for large vision language models, 2024.

\bibitem{jiang2024}
Xuhui Jiang, Yuxing Tian, Fengrui Hua, Chengjin Xu, Yuanzhuo Wang, and Jian Guo.
\newblock A survey on large language model hallucination via a creativity perspective, 2024.

\bibitem{li2023camel}
Guohao Li, Hasan Abed Al~Kader Hammoud, Hani Itani, Dmitrii Khizbullin, and Bernard Ghanem.
\newblock Camel: Communicative agents for "mind" exploration of large language model society, 2023.

\bibitem{li2023blip2}
Junnan Li, Dongxu Li, Silvio Savarese, and Steven Hoi.
\newblock Blip-2: Bootstrapping language-image pre-training with frozen image encoders and large language models, 2023.

\bibitem{li2023objecthallucination}
Yifan Li, Yifan Du, Kun Zhou, Jinpeng Wang, Wayne~Xin Zhao, and Ji-Rong Wen.
\newblock Evaluating object hallucination in large vision-language models, 2023.

\bibitem{li2023evaluating}
Yifan Li, Yifan Du, Kun Zhou, Jinpeng Wang, Wayne~Xin Zhao, and Ji-Rong Wen.
\newblock Evaluating object hallucination in large vision-language models.
\newblock {\em arXiv preprint arXiv:2305.10355}, 2023.

\bibitem{liang2024divergentthink}
Tian Liang, Zhiwei He, Wenxiang Jiao, Xing Wang, Yan Wang, Rui Wang, Yujiu Yang, Zhaopeng Tu, and Shuming Shi.
\newblock Encouraging divergent thinking in large language models through multi-agent debate, 2024.

\bibitem{lin2014microsoft}
Tsung-Yi Lin, Michael Maire, Serge Belongie, James Hays, Pietro Perona, Deva Ramanan, Piotr Doll{\'a}r, and C~Lawrence Zitnick.
\newblock Microsoft coco: Common objects in context.
\newblock In {\em Computer Vision--ECCV 2014: 13th European Conference, Zurich, Switzerland, September 6-12, 2014, Proceedings, Part V 13}, pages 740--755. Springer, 2014.

\bibitem{liu2024improved}
Haotian Liu, Chunyuan Li, Yuheng Li, and Yong~Jae Lee.
\newblock Improved baselines with visual instruction tuning.
\newblock In {\em Proceedings of the IEEE/CVF Conference on Computer Vision and Pattern Recognition}, pages 26296--26306, 2024.

\bibitem{CognitiveSynergy}
Andrea~I Luppi, Pedro~AM Mediano, Fernando~E Rosas, Negin Holland, Tim~D Fryer, John~T O’Brien, James Rowe, David~K Menon, Daniel Bor, and Emmanuel A~Stamatak is.
\newblock Asynergistic core for human brain evolution and cognition.
\newblock {\em Nature Neuroscience}, 25(6):771--782, 2022.

\bibitem{madaan2023}
Aman Madaan, Niket Tandon, Prakhar Gupta, Skyler Hallinan, Luyu Gao, Sarah Wiegreffe, Uri Alon, Nouha Dziri, Shrimai Prabhumoye, Yiming Yang, Shashank Gupta, Bodhisattwa~Prasad Majumder, Katherine Hermann, Sean Welleck, Amir Yazdanbakhsh, and Peter Clark.
\newblock Self-refine: Iterative refinement with self-feedback, 2023.

\bibitem{qiao2023reasoning}
Shuofei Qiao, Yixin Ou, Ningyu Zhang, Xiang Chen, Yunzhi Yao, Shumin Deng, Chuanqi Tan, Fei Huang, and Huajun Chen.
\newblock Reasoning with language model prompting: A survey, 2023.

\bibitem{schwenk2022okvqa}
Dustin Schwenk, Apoorv Khandelwal, Christopher Clark, Kenneth Marino, and Roozbeh Mottaghi.
\newblock A-okvqa: A benchmark for visual question answering using world knowledge.
\newblock In {\em European conference on computer vision}, pages 146--162. Springer, 2022.

\bibitem{shinn2023}
Noah Shinn, Federico Cassano, Edward Berman, Ashwin Gopinath, Karthik Narasimhan, and Shunyu Yao.
\newblock Reflexion: Language agents with verbal reinforcement learning, 2023.

\bibitem{sun2023}
Zhiqing Sun, Sheng Shen, Shengcao Cao, Haotian Liu, Chunyuan Li, Yikang Shen, Chuang Gan, Liang-Yan Gui, Yu-Xiong Wang, Yiming Yang, Kurt Keutzer, and Trevor Darrell.
\newblock Aligning large multimodal models with factually augmented rlhf, 2023.

\bibitem{team2023gemini}
Gemini Team, Rohan Anil, Sebastian Borgeaud, Yonghui Wu, Jean-Baptiste Alayrac, Jiahui Yu, Radu Soricut, Johan Schalkwyk, Andrew~M Dai, Anja Hauth, et~al.
\newblock Gemini: a family of highly capable multimodal models.
\newblock {\em arXiv preprint arXiv:2312.11805}, 2023.

\bibitem{wang2024vigc}
Bin Wang, Fan Wu, Xiao Han, Jiahui Peng, Huaping Zhong, Pan Zhang, Xiaoyi Dong, Weijia Li, Wei Li, Jiaqi Wang, and Conghui He.
\newblock Vigc: Visual instruction generation and correction, 2024.

\bibitem{wei2023chainofthought}
Jason Wei, Xuezhi Wang, Dale Schuurmans, Maarten Bosma, Brian Ichter, Fei Xia, Ed Chi, Quoc Le, and Denny Zhou.
\newblock Chain-of-thought prompting elicits reasoning in large language models, 2023.

\bibitem{collectiveintelligence}
Anita~Williams Woolley, Christopher~F Chabris, Alex Pentland, Nada Hashmi, and Thomas W~Mal one.
\newblock Evidence for a collective intelligence factor in the performance of human groups.
\newblock {\em science}, 330(6004):686--688, 2010.

\bibitem{yao2023tree}
Shunyu Yao, Dian Yu, Jeffrey Zhao, Izhak Shafran, Thomas~L. Griffiths, Yuan Cao, and Karthik Narasimhan.
\newblock Tree of thoughts: Deliberate problem solving with large language models, 2023.

\bibitem{yu2024hallucidoctor}
Qifan Yu, Juncheng Li, Longhui Wei, Liang Pang, Wentao Ye, Bosheng Qin, Siliang Tang, Qi Tian, and Yueting Zhuang.
\newblock Hallucidoctor: Mitigating hallucinatory toxicity in visual instruction data, 2024.

\bibitem{zhang2023benchmarking}
Tianyi Zhang, Faisal Ladhak, Esin Durmus, Percy Liang, Kathleen McKeown, and Tatsunori~B. Hashimoto.
\newblock Benchmarking large language models for news summarization, 2023.

\bibitem{zhao2024assessing}
Yunpu Zhao, Rui Zhang, Wenyi Li, Di Huang, Jiaming Guo, Shaohui Peng, Yifan Hao, Yuanbo Wen, Xing Hu, Zidong Du, Qi Guo, Ling Li, and Yunji Chen.
\newblock Assessing and understanding creativity in large language models, 2024.

\bibitem{zhu2023minigpt}
Deyao Zhu, Jun Chen, Xiaoqian Shen, Xiang Li, and Mohamed Elhoseiny.
\newblock Minigpt-4: Enhancing vision-language understanding with advanced large language models.
\newblock {\em arXiv preprint arXiv:2304.10592}, 2023.

\end{thebibliography}
}

\end{document}